\documentclass{article}
\newcommand{\comment}[1]{}
\usepackage[T1]{fontenc}
\usepackage[utf8x]{inputenc}
\usepackage [english] { babel }
\usepackage{hyperref}
\usepackage{amsthm}
\usepackage{algorithm2e}
\usepackage{mathtools}
\usepackage{graphicx}
\usepackage{amsmath,amssymb,amsfonts,mathrsfs}
\usepackage{enumitem}
\usepackage{tikz}
\usetikzlibrary{bayesnet}
\usepackage{tkz-graph}
\usepackage{diagbox}

\DeclareGraphicsExtensions{.eps,.pdf,.jpg,.gif}

\theoremstyle{definition}

\newcommand{\B}[1]{\boldsymbol #1}

\title{Integrating Learning and Reasoning with\\
Deep Logic Models}

\author{Giuseppe Marra$^{1,2}$, Francesco Giannini$^{2}$ \\
Michelangelo Diligenti$^{2}$ and Marco Gori$^{2}$ \\
$^{1}$Department of Information Engineering \\ University of Florence, ITALY\\
$^{2}$Department of Information Engineering and Mathematics\\
University of Siena, ITALY\\
e-mail: g.marra@unifi.it, \{fgiannini,diligmic,marco\}@diism.unisi.it
}

\begin{document}

\maketitle

\begin{abstract}
Deep learning is very effective at jointly learning feature representations and classification models, especially when dealing with high dimensional input patterns. Probabilistic logic reasoning, on the other hand, is capable to take consistent and robust decisions in complex environments. The integration of deep learning and logic reasoning is still an open-research problem and it is considered to be the key for the development of real intelligent agents. This paper presents Deep Logic Models, which are deep graphical models integrating deep learning and logic reasoning both for learning and inference. Deep Logic Models create an end-to-end differentiable architecture, where deep learners are embedded into a network implementing a continuous relaxation of the logic knowledge. The learning process allows to jointly learn the weights of the deep learners and the meta-parameters controlling the high-level reasoning. The experimental results show that the proposed methodology overtakes the limitations of the other approaches that have been proposed to bridge deep learning and reasoning. 
\end{abstract}

\section{Introduction}

Artificial Intelligence (AI) approaches can be generally divided into symbolic and sub-symbolic approaches. Sub-symbolic approaches like artificial neural networks have attracted most attention of the AI community in the last few years. Indeed, sub-symbolic approaches have got a large competitive advantage from the availability of a large amount of labeled data in some applications. In these contexts, sub-symbolic approaches and, in particular, deep learning ones are effective in processing low-level perception inputs~\cite{bengio2009learning,lecun1998gradient}. For instance, deep learning architectures have been achieved state-of-the-art results in a wide range of tasks, e.g. speech recognition, computer vision, natural language processing, where deep learning can effectively develop feature representations and classification models at the same time.

On the other hand, symbolic reasoning~\cite{DeRaedtProbLog2007,kimmig2012short,muggleton1994inductive}, which is typically based on logical and probabilistic inference, allows to perform high-level reasoning (possibly under uncertainty) without having to deal with thousands of learning hyper-parameters. Even if recent work has tried to gain insight on how a deep model works~\cite{mahendran2015understanding}, sub-symbolic approaches are still mostly seen as \emph{black-boxes}, whereas symbolic approaches are generally more easier to interpret, as the symbol manipulation or chain of reasoning can be unfolded to provide an understandable explanation to a human operator.

In spite of the incredible success of deep learning, many researchers have recently started to question the ability of deep learning to bring us real AI, because the amount and quality of training data would explode in order to jointly learn the high-level reasoning that is needed to perform complex tasks~\cite{battaglia2018relational}. For example, forcing some structure to the output of a deep learner has been shown to bring benefits in image segmentation tasks, even when simple output correlations were added to the enforced contextual information~\cite{chen2015learning}.

Blending symbolic and sub-symbolic approaches is one of the most challenging open problem in AI and, recently, a lot of works, often referred as neuro-symbolic approaches~\cite{garcez2012neural}, have been proposed by several authors~\cite{chen2015learning,hazan2016blending,manhaeve2018deepproblog,rocktaschel2016learning}.

In this paper, we present Deep Logic Models (DLMs), a unified framework to integrate logical reasoning and deep learning. DLMs bridge an input layer processing the sensorial input patterns, like images, video, text, from a higher level which enforces some structure to the model output. Unlike in Semantic-based Regularization~\cite{diligenti2017semantic} or Logic Tensor Networks~\cite{donadello2017logic}, the sensorial and reasoning layers can be jointly trained, so that the high-level weights imposing the output structure are jointly learned together with the neural network weights, processing the low-level input.
The bonding is very general as any (set of) deep learners can be integrated and any output structure can be expressed. This paper will mainly focus on expressing the high-level structure using logic formalism like first--order logic (FOL). In particular, a consistent and fully differentiable relaxation of FOL is used to map the knowledge into a set of potentials that can be used in training and inference.

The outline of the paper is the following. Section~\ref{sec:model} presents the model and the integration of logic and learning.
Section~\ref{sec:related_works} compares and connects the presented work with previous work in the literature and Section~\ref{sec:exp_results} shows the experimental evaluation of the proposed ideas on various datasets.
Finally, Section~\ref{sec:conclusions} draws some conclusions and highlights some planned future work.

\section{Model}
\label{sec:model}
We indicate as $\B \theta$ the model parameters, and $X$ the collection of input sensorial data.
Deep Logic Models (DLMs) assume that the prediction of the system is constrained by the available prior knowledge. Therefore, unlike standard Neural networks which compute the output via a simple forward pass, the output computation in  DLM can be decomposed into two stages: a \emph{low-level} stage processing the input patterns, and a subsequent \emph{semantic} stage, expressing constraints over the output and performing higher level reasoning.
We indicate by $\B y = \{ y_1, \ldots, y_n \}$ and by $\B f = \{ f_1, \ldots, f_n \}$
the two multivariate random variables corresponding to the output of the model and to the output of the first stage respectively, where $n>0$ denotes the dimension of the model outcomes.
Assuming that the input data is processed using neural networks, the model parameters can be split into two independent components $\B \theta = \{ \B w, \B \lambda \}$, where $\B w$ is the vector of weights of the networks $f_{nn}$ and $\B \lambda$ is the vector of weights of the second stage, controlling the semantic layer and the constraint enforcement.
\begin{figure}[t]
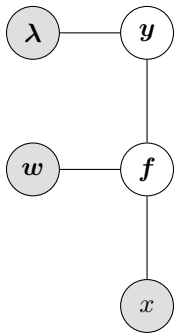

\centering
      \tikz{
        \node[obs] (x) {$x$} ;
        \node[latent, above=of x, yshift=0.1cm] (f) {$\B f$} ;
        \node[latent, above=of f, yshift=0.1cm] (y) {$\B y$} ;
        \node[obs, left=of f, xshift=0.2cm] (w) {$\B w$} ;
        \node[obs, left=of y, xshift=0.2cm] (lambda) {$\B \lambda$} ;
        \edge[-]{x} {f} ; 
        \edge[-]{f} {y} ; 
        \edge[-]{w} {f} ; 
        \edge[-]{lambda} {y} ; 
      }
    \caption{The DLM graphical model assumes that the output variables $\B y$ depend on the output of first stage $\B f$, processing the input X. This corresponds to the breakdown into a lower sensorial layer and a high level semantic one.}
    \label{fig:dlm_graphical_model}
\end{figure}
Figure \ref{fig:dlm_graphical_model} shows the graphical dependencies among the stochastic variables that are involved in our model. The first layer processes the inputs returning the values $\B f$ using a model with parameters $\B w$. The higher layer takes as input $\B f$ and applies reasoning using a set of constraints, whose parameters are indicated as $\B \lambda$, then it returns the set of output variables $\B y$.

The Bayes rule allows to link the probability of the parameters to the posterior and prior distributions:
\[
p(\B \theta | \B y, X) \propto p(\B y | \B \theta , X) p(\B \theta) \ .
\]
Assuming the breakdown into a sensorial and a semantic level, the prior may be decomposed as $p(\B \theta) = p(\B \lambda) p(\B w)$, while the posterior can be computed by marginalizing over the assignments for $\B f$:
\begin{equation}
    p(\B y | \B \theta , X) = \int_{\B f} p(\B y | \B f, \B \lambda) \cdot p(\B f | \B w, X) d\B f \ .
    \label{eq:posterior_exact}
\end{equation}
A typical choice is to link $p(\B f | \B w, X)$ to the outputs of the neural architectures:
\[
p(\B f | \B w, X) = \frac{1}{Z(\B f)}\exp\left( -\frac{(\B f-\B f_{nn})^2}{2\sigma^2}\right) \ ,
\]
where the actual (deterministic) output of the networks $f_{nn}$ over the inputs is indicated as $\B f_{nn}$. Please note that there is a one-to-one correspondence among each element of $\B y, \B f$ and $\B f_{nn}$, such that $|\B y|=|\B f|=|\B f_{nn}|$.

However, the integral in Equation~(\ref{eq:posterior_exact}) is too expensive to compute and, as commonly done in the deep learning community, only the actual output of the network is considered, namely:
\[
p(\B f | \B w, X) \approx \delta(\B f - \B f_{nn}) \ ,
\]
resulting in the following approximation of the posterior:
\[
p(\B y | \B \theta , X) \approx p(\B y | \B f_{nn}, \B \lambda) \ .
\]

A Deep Logic Model assumes that $p(\B y | \B f_{nn}, \B \lambda)$ is modeled via an undirected probabilistic graphical model in the exponential family, such that:
\begin{equation}
    \label{eq:model}
    p(\B y | \B f_{nn}, \B \lambda) \triangleq \frac{1}{Z(\B y)} \exp{\left(\Phi_r(\B y, \B f_{nn}) + \sum_c \lambda_c \Phi_c(\B y)\right)} \ ,
\end{equation}
where the $\Phi_c$ are potential functions expressing some constraints on the output variables, $\B \lambda= \{\lambda_1, \lambda_2, \ldots, \lambda_C \}$ are parameters controlling the confidence for the single constraints where a higher value corresponds to a stronger enforcement of the corresponding constraint, $\Phi_r$ is a potential favoring solutions where the output closely follows the predictions provided by the neural networks (for instance $\Phi_r(\B y,\B f_{nn})=-\frac{1}{2}||\B y- \B f_{nn}||^2$) and $Z(\B y)$ is a normalization factor (i.e. the partition function):
\[
Z(\B y) = \int_{{\B y}} \exp{\left(\Phi_r(\B y, \B f_{nn}) + \sum_c \lambda_c \Phi_c(\B y)\right)} d{\B y}.
\]

\subsection{MAP Inference}
\label{sec:map_inference}
MAP inference assumes that the model parameters are known and it aims at finding the assignment maximizing $p(\B y | \B f_{nn},\B \lambda)$. MAP inference does not require to compute the partition function $Z$ which acts as a constant when the weights are fixed. Therefore:
\[
{\B y}_M  = \text{arg}\!\max_{\B y} \log p({\B y} | \B f_{nn}, \B \lambda) =
\text{arg}\!\max_{\B y}\left[ \Phi_r(\B y, \B f_{nn}) + \sum_c \lambda_c \Phi_c(\B y)\right].
\]
The above maximization problem can be optimized via gradient descent by computing:
\[
\nabla_{\B y}\log p({\B y} | \B f_{nn}, \B \lambda) = \nabla_{\B y} \Phi_{r}(\B y, \B f_{nn}) +\sum_c \lambda_c \nabla_{\B y} \Phi_c(\B y) 
\]

\subsection{Learning}
\label{sec:learning}
Training can be carried out by maximizing the likelihood of the training data:
\[
\text{arg}\!\max_{\B \theta} \log p(\B \theta | \B y_t, X) =  \log p(\B y_t | \B \theta, X) + \log p(\B w) + \log p(\B \lambda) \ .
\]
In particular, assuming that $p(\B y_t | \B \theta, X)$ follows the model defined in equation (\ref{eq:model}) and the parameter priors follow Gaussian distributions, we get:
\[
\log p(\B \theta | \B y_t, X) \!=\!
-\frac{\alpha}{2} ||\B w||^2 -\frac{\beta}{2} ||\B \lambda||^2  -
\Phi_r(\B y_t, \B f_{nn}) +\sum_c \lambda_c \Phi_c(\B y_t) - \log Z(\B y)
\]
where $\alpha,\beta$ are meta-parameters determined by the variance of the selected Gaussian distributions.
Also in this case the likelihood may be maximized by gradient descent using the following derivatives with respect to the model parameters:
\[
\begin{array}{lcl}
\frac{\partial \log p(\B \theta | \B y_t, X)}{\partial \lambda_c} &=& -\beta \lambda_c + \Phi_c(\B y_t) - E_p\left[\Phi_c\right] \\
&& \\
\frac{\partial \log p(\B \theta | \B y_t, X)}{\partial w_i} &=&
-\alpha w_i + \frac{\partial \Phi_r(\B y_t,\B f_{nn})}{\partial w_i} - E_p\left[\frac{\partial\Phi_r}{\partial w_i}\right]
\end{array}
\]
Unfortunately, the direct computation of the expected values in the above derivatives is not feasible. A possible approximation~\cite{goodfellow2016deep,haykin1994} relies on replacing the expected values with the corresponding value at the MAP solution, assuming that most of the probability mass of the distribution is centered around it.
This can be done directly on the above expressions for the derivatives or in the log likelihood:
\[
\log p(\B y_t | \B f_{nn}, X) \approx
\Phi_r(\B y_t, \B f_{nn}) - \Phi_r(\B y_M, \B f_{nn}) + \sum_c \lambda_c \left(\Phi_c(\B y_t) - \Phi_c(\B y_M)\right)
\]
From the above approximation, it emerges that the likelihood tends to be maximized when the MAP solution is close to the training data, namely if $\Phi_r(\B y_t, \B f_{nn}) \simeq \Phi_r(\B y_{M}, \B f_{nn})$ and $\Phi_c(\B y_t) \simeq \Phi_c(\B y_{M}) ~ \forall c$.
Furthermore, the probability distribution is more centered around the MAP solution when $\Phi_r(\B y_{M}, \B f_{nn})$ is close to its maximum value.
We assume that $\Phi_r$ is negative and have zero as upper bound: $\Phi_r(\B y,\B f_{nn})\le 0 ~\forall \B y,\B f_{nn}$, like it holds for example for the already mentioned negative quadratic potential $\Phi_r(\B y,\B f_{nn})=-\frac{1}{2}||\B y- \B f_{nn}||^2$. Therefore, the constraint $\Phi_r(\B y_t, \B f_{nn}) \simeq \Phi_r(\B y_{M}, \B f_{nn})$ is transformed into the two separate constraints $\Phi_r(\B y_t, \B f_{nn}) \simeq 0$ and $\Phi_r(\B y_{M}, \B f_{nn}) \simeq 0$.

This means that, given the current MAP solution, it is possible to increase the log likelihood by computing the gradient and weight updates using the following cost function:
\[
\begin{array}{c}
\log p(\B w) + \log p(\B \lambda) + \Phi_r(\B y_t, \B f_{nn}) + \Phi_r(\B y_{M}, \B f_{nn}) + \displaystyle\sum_c \lambda_c \left[\Phi_c(\B y_t) - \Phi_c(\B y_{M})\right]
\end{array}
\]
In this paper, a quadratic form for the priors and the potentials is selected, but other choices are possible. For example, $\Phi_r(\cdot)$ could instead be implemented as a negative cross entropy loss. Therefore, replacing the selected forms for the potentials and changing the sign to transform a maximization into a minimization problem, yields the following cost function, given the current MAP solution:
\begin{eqnarray}
C_{\B \theta}(\B y_t, \B y_{M} ,X) &=&\frac{\alpha}{2} ||\B w||^2 +\frac{\beta}{2} ||\B \lambda||^2 +\frac{1}{2}||\B y_t-\B f_{nn}||^2 + \frac{1}{2}||\B y_{M}-\B f_{nn}||^2 + \nonumber\\
&+&\displaystyle\sum_c \lambda_c \left[ \Phi_c(\B y_t) - \Phi_c(\B y_{M}) \right] \ .
\label{eq:local_cost_function}
\end{eqnarray}

Minimizing the cost function $C_{\B \theta}(\B y_t, \B y_{M} ,X)$ is just a local approximation of the full likelihood maximization for the current MAP solution. Therefore, the training process alternates the computation of the MAP solution, the computation of the gradient for $C_{\B \theta}(\B y_t, \B y_{M} ,X)$ and one weight update step.
\begin{algorithm}[t]
 \KwData{Input data $X$, output targets $\B y_t$, function models with weights $\B w$}
 \KwResult{Trained model parameters $\B \theta = \{\B \lambda, \B w\}$}
 Initialize $i=0$, $\B \lambda=\B 0$, random $\B w$;\\
 \While{not converged $\land ~ i<max\_iterations$}{
  Compute function outputs $\B f_{nn}$ on $X$ using current function weights $\B w$; \\
  Compute MAP solution ${\B y}_M  = \text{arg}\!\max_{\B y} \log p({\B y} | \B f_{nn}, \B \lambda)$;\\
  Compute gradient $\nabla_{\B \theta} C_{\B \theta}(\B y_t, \B y_{M} ,X)$;\\
  Update $\B \theta$ via gradient descent: $\B \theta_{i+1} = \B \theta_i - \lambda_{lr} \cdot \nabla_{\B \theta} C_{\B \theta}(\B y_t, \B y_{M} ,X)$;\\
  Set i=i+1;
 }
 \caption{Iterative algorithm to train the function weights $\B w$ and the constraint weights $\B \lambda$.}
 \label{al:training}
\end{algorithm}
Algorithm~\ref{al:training} summarizes this iterative training algorithm. Please note that, for any constraint $c$, the parameter $\lambda_c$ admits also a negative value. This is in case the $c$-th constraint turns out to be too satisfied by the actual MAP solution with respect to the satisfaction degree on the training data.

\subsection{Mapping Constraints into a Continuous Logic}
\label{sec:logic_and_constraints}
The DLM model is absolutely general in terms of the constraints that can be expressed on the outputs. However, this paper mainly focuses on constraints expressed in the output space $\B y$ by means of first--order logic formulas.
Therefore, this section focuses on defining a methodology to integrate prior knowledge expressed via FOL into a continuous optimization process.

In this framework we only deal with closed FOL formulas, namely formulas where any variable occurring in predicates is quantified. In the following, given an $m$-ary predicate $p$ and a tuple  $(a_1,\ldots,a_m)\in Dom(p)$, we say that $p(a_1,\ldots,a_m)\in[0,1]$ is a  \emph{grounding} of $p$.
Given a grounding of the variables occurring in a FOL formula (namely a grounding for all the predicates involved in the formula), the truth degree of the formula for that grounding is computed using the t-norm fuzzy logic theory as proposed in~\cite{novak2012mathematical}.
The overall degree of satisfaction of a FOL formula is obtained by grounding all the variables in such formula and aggregating the values with different operators depending on the occurring quantifiers. The details of this process are explained in the following of the section.
\paragraph{Grounded Expressions.} Any fully grounded FOL rule corresponds to an expression in propositional logic and we start showing how a propositional logic expression may be converted into a differentiable form.
In particular, one expression tree is built for each considered grounded FOL rule, where any occurrence of the basic logic connectives ($\lnot,\land,\lor,\Rightarrow$) is replaced by a unit computing its corresponding fuzzy logic operation according to a certain logic semantics.  In this regard, some recent work shows how to get convex (or even linear) functional constraints exploiting the convex \L ukasiewicz fragment~\cite{giannini2018convex}.
The expression tree can take as input the output values of the grounded predicates and then recursively compute the output values of all the nodes in the expression tree. The value obtained on the root node is the result of the evaluation of the expression given the input grounded predicates.

\begin{table}[!t]
	\centering
    \begin{tabular}{|c|c|c|c|}
                \hline
                \diagbox{operation}{t-norm} & Product & Minimum & \L ukasiewicz \\
                \hline
                $a \land b$ & $a \cdot b$ & $\min(a, b$) & $\max(0, a+b-1$) \\
                \hline
                $a \lor b$ & $a + b - a \cdot b$ & $\max(a, b)$ & $\min(1, a + b)$ \\
                \hline
                $\lnot a$ & $1 - a$ & $1 - a$ & $1 - a$ \\
                \hline
                $a \Rightarrow b$ & $\min(1, \frac{b}{a})$ & $a\leq b?1:b$ & $\min(1,1-a+b)$ \\
                \hline
	\end{tabular}
    \caption{The Operations performed by the single units of an expression tree depending on the inputs $a,b$ and the used t-norm.}
    \label{tab:forward_operations}
\end{table}

Table~\ref{tab:forward_operations} shows the algebraic operations corresponding to the logic operators for different selections of the t-norms.
Please note that the logic operators are always monotonic with respect of any single variable, but they are not always differentiable (nor even continuous). However, the sub-space where the operators are non-differentiable has null-Lebesgue measure, therefore they do not pose any practical issue, when used as part of a gradient descent optimization schema as detailed in the following.

We assume that the input data $X$ can be divided into a set of sub-domains $X=\{X_1,X_2, \ldots \}$, such that each variable $v_i$ of a FOL formula ranges over the data of one input domain, namely $v_i \in X_{d_i}$, where $d_i$ is the index of the domain for the variable $v_i$.

For example, let us consider the rule $\forall v_1 \forall v_2~ \lnot A(v_1,v_2) \land B(v_1)$. For any assignment to $v_1$ and $v_2$, the expression tree returns the output value $[1 - A(v_1,v_2)] \cdot B(v_1)$, assuming to exploit the product t-norm to convert the connectives.

\paragraph{Quantifiers.} The truth degree of a formula containing an expression with a universally quantified variable $v_i$ is computed as the average of the t-norm truth degree of the expression, when grounding $v_i$ over its domain.
The truth degree of the existential quantifier is the \textit{maximum} of the t-norm expression grounded over the domain of the quantified variable.
When multiple quantified variables are present, the conversion is performed from the outer to the inner variable. When only universal quantifiers are present the aggregation is equivalent to the overall average over each grounding.

In the previous example, this yields the following expression:
\begin{equation}
\label{eq:potential_example}
\Phi(X, A, B) = \frac{1}{|X_{d_1}| \cdot |X_{d_2}|}
\displaystyle\sum_{v_1 \in X_{d_1}} \displaystyle\sum_{v_2 \in X_{d_2}} [1 - A(v_1, v_2)] \cdot B(v_1) \ .
\end{equation}

\subsection{Potentials expressing the logic knowledge}
It is now possible to explain how to build the potentials from the prior knowledge.
In any learning task, each unknown grounded predicate corresponds to one variable in the vector $\B y$. In the above example, the number of groundings is $|X_{d_1}|\times |X_{d_2}|$ (i.e. the size of the cartesian product of the domains of $A$) and $|X_{d_1}|$ (i.e. the size of the domain of $B$). Therefore, assuming that both predicates $A,B$ are unknown, $|\B y|=|\B f|=|X_{d_1}|\times |X_{d_2}| + |X_{d_1}$|. The vector $\B f_{nn}$ is built similarly by replacing a generic predicate with its neural implementation and then emplacing the function values for the groundings in the vector. Again for the considered example: 
$\B f_{nn} = \{ f_A(v_{11},v_{21}), \ldots , f_A(v_{1|X_{d_1}|},v_{2|X_{d_2}|}), f_B(v_{11}), \ldots , f_B(v_{1|d_2|}) \}$, where $v_{ij}$ is the $j$-th grounding for the $i$-th variable and $f_A,f_B$ are the learned neural approximations of $A$ and $B$, respectively.
Finally, the differentiable potential for the example formula is obtained by replacing in Equation~(\ref{eq:potential_example}) each grounded predicate with the corresponding stochastic variable in $\B y$.

\begin{figure}[t]
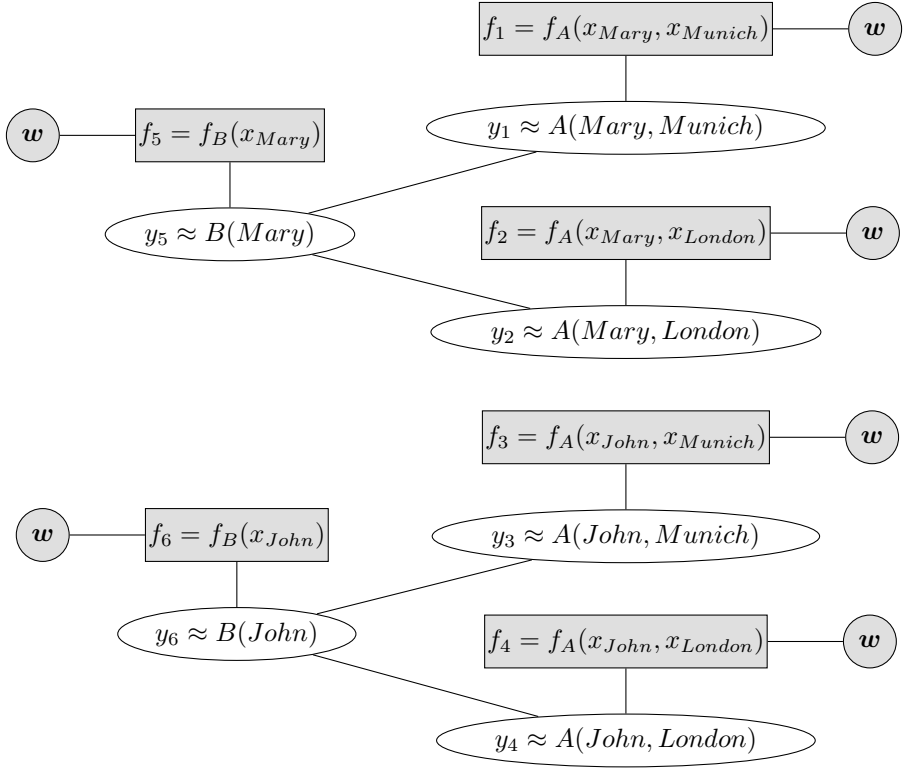

\centering
      \tikz{
        \node[latent, ellipse,draw] (y1) {$y_1 \approx  A(Mary,Munich)$} ;
        \node[obs, rectangle,draw, above=of y1, yshift=-0.4cm] (f1) {$f_1 = f_A(x_{Mary}, x_{Munich})$} ;
        \edge[-]{y1} {f1} ;
        \node[obs, right=of f1] (w1) {$\B w$} ;
        \edge[-]{w1} {f1} ;

        \node[latent, ellipse,draw, below=of y1, yshift=-1.0cm] (y2) {$y_2 \approx  A(Mary,London)$} ;
        \node[obs, rectangle,draw, above=of y2, yshift=-0.4cm] (f2) {$f_2 = f_A(x_{Mary}, x_{London})$} ;
        \edge[-]{y2} {f2} ;
        \node[obs, right=of f2] (w2) {$\B w$} ;
        \edge[-]{w2} {f2} ;

        \node[latent, ellipse,draw, below=of y2, yshift=-1.0cm] (y3) {$y_3 \approx  A(John,Munich)$} ;
        \node[obs, rectangle,draw, above=of y3, yshift=-0.4cm] (f3) {$f_3 = f_A(x_{John}, x_{Munich})$} ;
        \edge[-]{y3} {f3} ;
        \node[obs, right=of f3] (w3) {$\B w$} ;
        \edge[-]{w3} {f3} ;

        \node[latent, ellipse,draw, below=of y3, yshift=-1.0cm] (y4) {$y_4 \approx  A(John,London)$} ;
        \node[obs, rectangle,draw, above=of y4, yshift=-0.4cm] (f4) {$f_4 = f_A(x_{John}, x_{London})$} ;
        \edge[-]{y4} {f4} ;
        \node[obs, right=of f4] (w4) {$\B w$} ;
        \edge[-]{w4} {f4} ;

        \node[latent, ellipse,draw, left=of y2, yshift=+1.3cm] (y5) {$y_5 \approx  B(Mary)$} ;
        \node[obs, rectangle,draw, above=of y5, yshift=-0.4cm] (f5) {$f_5 = f_B(x_{Mary})$} ;
        \edge[-]{y5} {f5} ;
        \edge[-]{y5} {y1} ; 
        \edge[-]{y5} {y2} ;
        \node[obs, left=of f5] (w5) {$\B w$} ;
        \edge[-]{w5} {f5} ;

        \node[latent, ellipse,draw, left=of y3, yshift=-1.3cm] (y6) {$y_6 \approx  B(John)$} ;
        \node[obs, rectangle,draw, above=of y6, yshift=-0.4cm] (f6) {$f_6 = f_B(x_{John})$} ;
        \edge[-]{y6} {f6} ;
        \edge[-]{y6} {y3} ; 
        \edge[-]{y6} {y4}; 
        \node[obs, left=of f6] (w6) {$\B w$} ;
        \edge[-]{w6} {f6} ;
      }
    \caption{The undirected graphical model built by a DLM for the rule $\forall v_1 \forall v_2~ \lnot A(v_1,v_2) \land B(v_1)$ where $v_1$ can assume values over the constants $\{Mary, John\}$ and $v_2$ over $\{ Munich, London\}$. Each stochastic node $y_i$ approximates one grounded predicate, while the $f_i$ nodes are the actual output of a network getting the pattern representations of a grounding.}
    \label{fig:logic_graphical_model}
\end{figure}

Figure~\ref{fig:logic_graphical_model} shows the undirected graphical model corresponding to the DLM for the running example rule used in this section, assuming that $v_1$ can assume values over the constants $\{Mary, John\}$ and $v_2$ over $\{ Munich, London\}$. Each stochastic node $y_i$ approximates one grounded predicate, while the $f_i$ nodes are the actual output of a neural network getting as input the pattern representations of the corresponding grounding. The vertical connections between two $y_i$ and $f_i$ nodes correspond to the cliques over the groundings for which the $\Phi_r$ potential can be decomposed. The links between the $y_i$ nodes corresponds to the cliques over the groundings of the rule for which the corresponding $\Phi_c$ potential can be decomposed. The structure of these latter cliques follows a template determined by the rule, that is repeated for the single groundings. The graphical model is similar to the ones built by Probabilistic Soft Logic~\cite{bach2015hinge} or Markov Logic Networks~\cite{richardson2006markov}, but enriched with the nodes corresponding to the output of the neural networks.

\section{Related Works}
\label{sec:related_works}
DLMs have also their roots in Probabilistic Soft Logic (PSL)~\cite{bach2015hinge}, a probabilistic logic using an undirected graphical model to represent a grounded FOL knowledge base, and employing a similar differentiable and convex approximation of FOL. PSL, similar to a DLM, allows to learn the weight of each formula in the KB by maximizing the log likelihood of the training data. However in PSL, rule weights are restricted to only positive values denoting how far the rule is from being satisfied. On the other hand, in DLMs the rule weights denote the needed constraint reactions to match the degree satisfaction of the training data. In addition, unlike DLMs, PSL focuses on logic reasoning without any integration with deep learners, beside a simple stacking with no joint training.

The integration of learning from data and symbolic reasoning~\cite{garcez2012neural} has recently attracted a lot of attention. Hu at al.~\cite{hu2016harnessing}, Semantic-based regularization (SBR)~\cite{diligenti2017semantic} for kernel machines and Logic Tensor Networks (LTN)~\cite{donadello2017logic} for neural networks share the same basic idea of integrating logic reasoning and learning using a similar continuous relaxation of logic to the one presented in this paper. However, this class of approaches considers the reasoning layer as frozen, without allowing to jointly train its parameters. This is a big limitation, as these methods work better only with hard constraints, while they are less suitable in presence of reasoning under uncertainty.

The integration of deep learning with Conditional Random Fields (CRFs)~\cite{Xuezhe2016} is also an alternative approach to enforce some structure on the network output. This approach has been proved to be quite successful on sequence labeling for natural language processing tasks. This methodology can be seen as a special case of the more general methodology presented in this paper, when the potential functions are used to represent the correlations among consecutive outputs of a recurrent deep network.

DeepProbLog~\cite{manhaeve2018deepproblog} extends the popular ProbLog~\cite{DeRaedtProbLog2007} probabilistic programming framework with the integration of deep learners.
DeepProbLog requires the output from the neural networks to be probabilities and an independence assumption among atoms in the logic is required to make inference tractable. This is a strong restriction, since the sub-symbolic layer often consists of several neural layers sharing weights.

A Neural Theorem Prover (NTP)~\cite{rocktaschel2016learning, rocktaschel2017end} is an end-to-end differentiable prover based on the Prolog’s backward chaining algorithm. An NTP constructs an end-to-end differentiable architecture capable of proving queries to a KB using sub-symbolic vector representations. NTPs have been proven to be effective in tasks like entity linking and knowledge base completion. However, an NTP encodes relations as vectors using a frozen pre-selected function (like cosine similarity). This can be ineffective in modeling relations with a complex and multifaceted nature (for example a relation \texttt{friend(A,B)} can be triggered by different relationships of the representations in the embedding space). On the other hand, DLMs allow a relation to be encoded by any selected function (e.g. any deep neural networks), which is co-trained during learning. Therefore, DLMs are capable of a more powerful and flexible exploitation of the representation space. On the other end, DLMs require to fully ground a KB (like SBR, LTN, PSL and most of other methods discussed here), while NTPs expands only the groundings on the explored frontier, which can be more efficient in some cases.

Deep Structured Models~\cite{chen2015learning,lin2016efficient} use a similar graphical model to bridge the sensorial and semantic levels. However, they have mainly focused on imposing correlations on the output layer, without any focus on logic reasoning. Furthermore, DLMs transform the training process into an iterative constrained optimization problem, which is very different from the approximation of the partition function used in Deep Structured Models.

DLMs also open up the possibility to iteratively integrate rule induction mechanisms like the ones proposed by the Inductive Logic Programming community~\cite{lavrac1994inductive,quinlan1990learning}.

\section{Experimental Results}
\label{sec:exp_results}
\subsection{The PAIRS artificial dataset}
\begin{figure}
	\centering
	\includegraphics[width=1\textwidth]{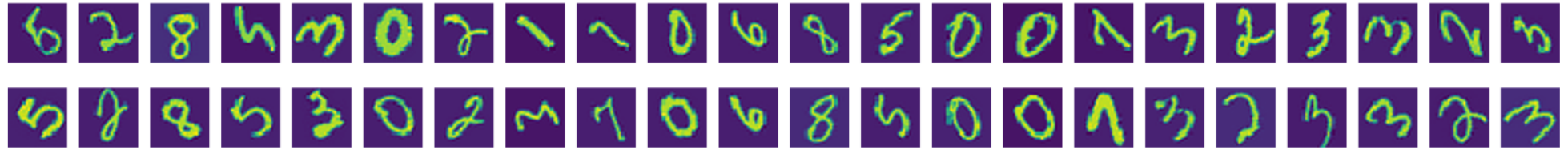}
	\caption{A sample of the data used in the PAIRS experiment, where each column is a pair of digits.}
	\label{fig:pairs}
\end{figure}

Consider the following artificial task. We are provided with 1000 pairs of handwritten digits images sampled from the MNIST dataset. The pairs are not constructed randomly but they are compilied according to the following structure:
\begin{enumerate}
    \item pairs with mixed even-odd digits are not allowed;
    \item the first image of a pair represents a digit randomly selected from a uniform distribution;
    \item if the first image is an even (resp. odd) digit, the second image of a pair represents one of the five even (resp. odd) digits with probabilities $p_1 \ge  p_2 \ge p_3 \ge p_4 \ge p_5$, with $p_1$ the probability of being an image of the same digit, $p_2$ the probability of being an image of the next even/odd digit, and so on.
\end{enumerate}
For example, if the first image of a pair is selected to be a \textit{two}, the second image will be a \textit{two} with probability $p_1$, it will be a \textit{four} with probability $p_2$, a \textit{six} with probability $p_3$ and so on, in a circular fashion.
An example is shown in Figure \ref{fig:pairs}.
A correct classification happens when both digit in a pair are correctly predicted.

To model a task using DLMs there are some common design choices regarding these two features that one needs to take. We use the current example to show them.
The first choice is to individuate the \emph{constants} of the problem and their sensorial representation in the perceptual space.
Depending on the problem, the constants can live in a single or multiple separate domains. In the pairs example, the images are constants and each one is represented as a vector of pixel brightnesses like commonly done in deep learning. 

The second choice is the selection of the \emph{predicates} that should predict some characteristic over the constants and their implementation.
In the pairs experiment, the predicates are the membership functions for single digits (e.g. \texttt{one(x)}, \texttt{two(x)}, etc.). A single neural network with 1 hidden layer, 10 hidden neurons and $10$ outputs, each one mapped to a predicate, was used in this toy experiment.
The choice of a small neural network is due to the fact that the goal is not to get the best possible results, but to show how the prior knowledge can help a classifier to improve its decision. In more complex experiments, different networks can be used for different sets of predicates, or each use a separate network for each predicate.

Finally, the \emph{prior knowledge} is selected. In the pairs dataset, where the constants are grouped in pairs, it is natural to express the correlations among two images in a pair via the prior knowledge. Therefore, the knowledge consists of $100$ rules in the form $\forall (x,y)~ D_1(x) \rightarrow D_2(y)$, where $(x,y)$ is a generic pair of images and $(D_1,D_2)$ range over all the possible pairs of digit classes.

We performed the experiments with $p_1=0.9, p_2=0.07, p_3=p_4=p_5=0.01$. All the images are rotated with a random degree between $0$ and $90$ anti-clockwise to increase the complexity of the task. There is a strong regularity in having two images representing the same digit in a pair, even some rare deviations from this rule are possible. Moreover, there are some inadmissible pairs, i.e. those containing mixed even-odd digits. The train and test sets are built by sampling $90\%$ and $10\%$ image pairs.

The results provided using a DLM have been compared against the following baselines:
\begin{itemize}
	\item the neural network (NN) with no knowledge of the structure of the problem;
	\item the Semantic Based Regularization \cite{diligenti2017semantic} (SBR) framework, which also employs logical rules to improve the learner. However, the rule weights are treated as fixed parameters, which are not jointly trained during learning. Since searching in the space of these parameters via cross-validation is not feasible, a strong prior was provided to make SBR prefers pairs with the same image using 10 rules of the form $\forall (x,y)~ D(x) \rightarrow D(y)$, for each digit class $D$. These rules hold true in most cases and improve the baseline performance of the network. 
\end{itemize}

Table~\ref{tab:pairs_results} shows how the neural network output of a DLM (DLM-NN) already beats both the same neural model trained without prior knowledge and SBR. This happens because the neural network in DLM is indirectly adjusted to respect the prior knowledge in the overall optimization problem. When reading the DLM output from the MAP solution (DLM), the results are significantly improved.
\begin{table}[t]
\centering
	\label{tab:pairs_results}
	\centering
	\begin{tabular}{l|cccc}
		Model & NN & SBR & DLM-NN & DLM \\
		\hline
		Accuracy & 0.62 & 0.64  & 0.65   & 0.76 \\
	\end{tabular}
    \caption{Comparison of the accuracy metric on the PAIRS dataset using different models.}
\end{table}

\subsection{Link Prediction in Knowledge Graphs}
Neural-symbolic approaches have been proved to be very powerful to perform approximated logical reasoning~\cite{trouillon2016complex}. A common approach is to assign to each logical constant and relation a learned vectorial representation~\cite{bordes2013translating}. Approximate reasoning is then carried out in this embedded space. Link Prediction in Knowledge Graphs is a generic reasoning task where it is requested to establish the links of the graph between semantic entities acting as constants. Rocktaschel et al.~\cite{rocktaschel2017end} shows state-of-the-art performances on some link prediction benchmarks by combining Prolog backward chain with a soft unification scheme. 

This section shows how to model a link prediction task on the \emph{Countries} dataset using a Deep Logic Models, and compare this proposed solution to the other state-of-the-art approaches.

\paragraph{Dataset.} The Countries dataset~\cite{bouchard2015approximate} consists of $244$ countries (e.g. germany), $5$ regions (e.g. europe), $23$ sub-regions (e.g. western europe, northern america, etc.), which act as the constants of the KB. Two types of binary relations among the constant are present in the dataset: $\texttt{locatedIn}(c_1,c_2)$, expressing that $c_1$ is part of $c_2$  and $\texttt{neighborOf}(c_1,c_2)$, expressing that $c_1$ neighbors with $c_2$. The knowledge base consists of $1158$ facts about the countries, regions and sub-regions, expressed in the form of Prolog facts (e.g. \texttt{locatedIn}(italy,europe)).
The training, validation and test sets are composed by $204,20$ and $20$ countries, respectively, such that each country in the validation and test sets has at least one neighbor in the training set.
Three different tasks have been proposed for this dataset with an increasing level of difficulty. For all tasks, the goal is to predict the relation $\texttt{locatedIn}(c, r)$ for every test country $c$ and all five regions $r$, but the access to training atoms in the KB varies, as explained in the following:
\begin{itemize}
    \item Task S1: all ground atoms $\texttt{locatedIn}(c, r)$ where $c$ is a test country and $r$ is a region are removed from the KB. Since information about the sub-region of test countries is still contained in the KB, this task can be solved exactly by learning the transitivity of the \texttt{locatedIn} relation.
    \item Task S2: like S1 but all grounded atoms $\texttt{locatedIn}(c, s)$, where $c$ is a test country and $s$ is a sub-region,  are removed. The location of test countries needs to be inferred from the location of its neighbors. This task is more difficult than S1, as neighboring countries might not be in the same region.
    \item Task S3: like S2, but all ground atoms $\texttt{locatedIn}(c, r)$, where $r$ is a region and $c$ is a training country with either a test or validation country as a neighbor, are removed. This task requires multiple reasoning steps to determine an unknown link, and it strongly exploits the sub-symbolic reasoning capability of the model to be effectively solved.
\end{itemize}

\paragraph{Model.} Each country, region and sub-region corresponds to a constant. Since the constants are just symbols, each one is assigned to an embedding, which is learned together with the other parameters of the model.
The predicates are the binary relations \texttt{locatedIn} and \texttt{neighborOf}, which connect constants in the KB. Each relation is learned via a separate neural network with a $50$ neuron hidden layer taking as input the concatenation of the embeddings of the constants.
In particular, similarly to~\cite{bordes2013translating}, the constants are encoded into a one-hot vector, which is processed by the first layer of the network, outputting an embedding composed by $50$ real number values.
As commonly done in link prediction tasks, the learning process is performed in a transductive mode. In particular, the input $X$ consists of all possible constants for the task, while the train examples $\B y_t$ will cover only a subset of all the possible grounded predicates, leaving to the joint train and inference process the generalization of the prediction to the other unknown grounded relations. Indeed, the output of the train process in this case is both the set of model parameters and the MAP solution predicting the unknown grounded relations that hold true.

Multi-step dependencies among the constants are very important to predict the existence of a link in this task. For example in task S1, the prediction of a link among a country and a region can be established via the path passing by a sub-region, once the model learns a rule stating the transitivity of the \texttt{locatedIn} relation (i.e. $\texttt{locatedIn}(x,y) \land \texttt{locatedIn}(y,z) \rightarrow \texttt{locatedIn}(x,z)$). Exploiting instead the rule $\texttt{neighborOf}(x,y) \land \texttt{locatedIn}(y,z) \rightarrow \texttt{locatedIn}(x,z)$, the model should be capable of approximately solving task S2.

All $8$ rules $\forall x~\forall y~\forall z~\texttt{A}(x,y) \land \texttt{B(y,z)} \rightarrow \texttt{C}(y,z)$, where \texttt{A}, \texttt{B} and \texttt{C} are either \texttt{neighborOf} or \texttt{locatedIn} are added to the knowledge base for this experiment. These rules represent all the 2-steps paths reasoning that can be encoded, and the strength of each rule needs to be estimated as part of the learning process for each task.
The training process will iteratively minimize Equation~\ref{eq:local_cost_function} by jointly determining the embeddings and the network weights such that network outputs and the MAP solution will correctly predict the training data, while respecting the constraints on the MAP solution at the same level as on the train data.

\paragraph{Results.} Table~\ref{tab:countries} compares DLM against the state-of-the-art methods used by Rocktaschel et al.~\cite{rocktaschel2017end}, namely ComplEx, NTP and NTP$\lambda$.
Task S1 is the only one that can be solved exactly when the transitive property of the $\texttt{locatedIn}$ relation has been learned to always hold true. Indeed, most methods are able to perfectly solve this task, except for the plain NTP model.
DLM is capable perfectly solving this task by joining the logical reasoning capabilities with the discriminative power of neural networks. DLMs perform better than the competitors on tasks S2 and S3, thanks to additional flexibility obtained by jointly training the relation functions using neural networks, unlike the simple vectorial operations like the cosine similarity employed by the competitors.
\begin{table}[t]
\centering
	\label{tab:countries}
	\centering
	\begin{tabular}{l|cccc}
	\textbf{Task} & ComplEx & NTP & NTP$\lambda$ & DLM\\
	\hline
	     S1  & $99.37$ & $90.83$  & $\textbf{100.00}$ & $\textbf{100.00}$\\
	     \hline
         S2  & $87.95$ & $87.40$  & $93.04$ & $\textbf{97.79}$\\
         \hline
         S3  & $48.44$ & $56.68$  & $77.26$ & $\textbf{91.93}$\\
	\end{tabular}
    \caption{Comparison of the accuracy provided by different methods on link prediction on the Countries dataset. Bold numbers are the best performers for each task.}
\end{table}

\section{Conclusions and future work}
\label{sec:conclusions}
This paper presents Deep Logic Models that integrate (deep) learning and logic reasoning into a single fully differentiable architecture. The logic can be expressed with unrestricted FOL formalism, where each FOL rule is converted into a differentiable potential function, which can be integrated into the learning process. The main advantage of the presented framework is the ability to fully integrate learning from low-level representations and semantic high-level reasoning over the network outputs. Allowing to jointly learn the weights of the deep learners and the parameters controlling the reasoning enables a positive feedback loop, which is shown to improve the accuracy of both layers. Future work will try to bridge the gap between fully grounded methodologies like current Deep Logic Models and Theorem Provers which expand only the groundings needed to expand the frontier of the search space.

\bibliography{references}
\bibliographystyle{splncs03}

\end{document}